\documentclass[letterpaper, 10 pt, conference]{ieeeconf}  

\IEEEoverridecommandlockouts                              

\usepackage{multicol}
\usepackage{graphicx}
\usepackage[font=footnotesize,labelformat=simple]{subcaption}
\usepackage[font=footnotesize]{caption}
\usepackage{tikz}
\usepackage{tikzscale}
\usepackage{blindtext}
\usepackage{tikz-qtree}
\usepackage{float}
\usepackage{calc}
\usepackage{amsmath}
\usepackage{booktabs}
\usepackage{array}
\usepackage{wrapfig}
\usepackage{url}
\ifx\hidenotes\undefined
  \usepackage{xcolor}
  \newcommand{\xxnote}[3]{\color{#2}{#1: #3}}
\fi

\title{\LARGE \bf
Zero-Shot Imitating Collaborative Manipulation Plans \\ from YouTube Cooking Videos
}

\author{Hejia Zhang, Jie Zhong, Stefanos Nikolaidis
\thanks{Hejia Zhang, Jie Zhong and Stefanos Nikolaidis are with the Department of Computer Science, University of Southern California, Los Angeles, USA {\tt\small \{hejiazha,jzhong54, nikolaid\}@usc.edu }}%
}

\begin{document}

\maketitle
\thispagestyle{empty}
\pagestyle{empty}

\begin{abstract}

People often watch videos on the web to learn how to cook new recipes, assemble furniture or repair a computer. We wish to enable robots with the very same capability. This is challenging; there is a large variation in manipulation actions and some videos even involve multiple persons, who collaborate by sharing and exchanging objects and tools. Furthermore, the learned representations need to be general enough to be transferable to robotic systems. On the other hand, previous work has shown that the space of human manipulation actions has a linguistic, hierarchical structure that relates actions to manipulated objects and tools.  
Building upon this theory of language for action, we propose a system for understanding and executing demonstrated action sequences from full-length, real-world cooking videos on the web. The system takes as input a new, previously unseen cooking video annotated with object labels and bounding boxes, and outputs a collaborative manipulation action plan for one or more robotic arms. We demonstrate performance of the system in a standardized dataset of 100 YouTube cooking videos, as well as in six full-length Youtube videos that include collaborative actions between two participants. We compare our system with a baseline system that consists of a state-of-the-art action detection baseline and show our system achieves higher action detection accuracy. We additionally propose an open-source platform for executing the learned plans in a simulation environment as well as with an actual robotic arm. \footnote{Code of the proposed system is available at \url{https://bit.ly/3v524w6}.}

\end{abstract}

\section{INTRODUCTION}

We focus on the problem of learning collaborative manipulation plans for a robot. Our goal is to have the robot ``watch'' unconstrained videos on the web, extract the action sequences shown in the videos and convert them to an executable plan that it can perform either independently, or as part of a human-robot or robot-robot team. 

Learning from online videos is hard, particularly in collaborative settings: it requires recognizing the actions executed, together with manipulated tools and objects. In many collaborative tasks these actions include handing objects over or holding an object for the other person to manipulate. There is a very large variation in how the actions are performed and collaborative actions may overlap spatially and temporally~\cite{hoffman2019evaluating}.

On the other hand, actions exhibit a syntactic-like structure that is common in language. Yang and Aloimonos~\cite{yang2014cognitive,yang2015robot} build upon this theoretical insight to propose a manipulation action grammar, where symbolic information of manipulated objects and tools are parsed to interpret action sequences from 12 YouTube video clips.

\begin{figure}[!t]
\centering
\includegraphics[width=0.7\linewidth]{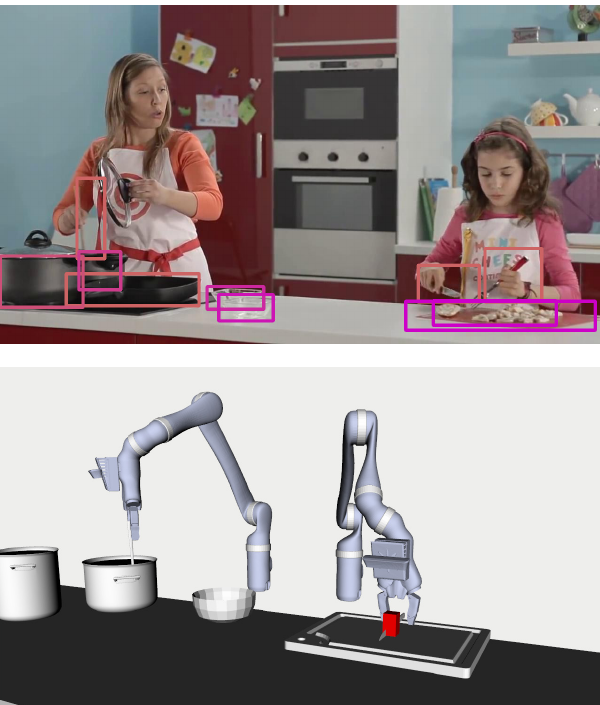}
\caption{The robots execute the action sequence shown in the video.}
\label{fig:best}
\vspace{-15pt}
\end{figure}

\begin{figure*}[!t]
\centering
\includegraphics[width=1.0\linewidth]{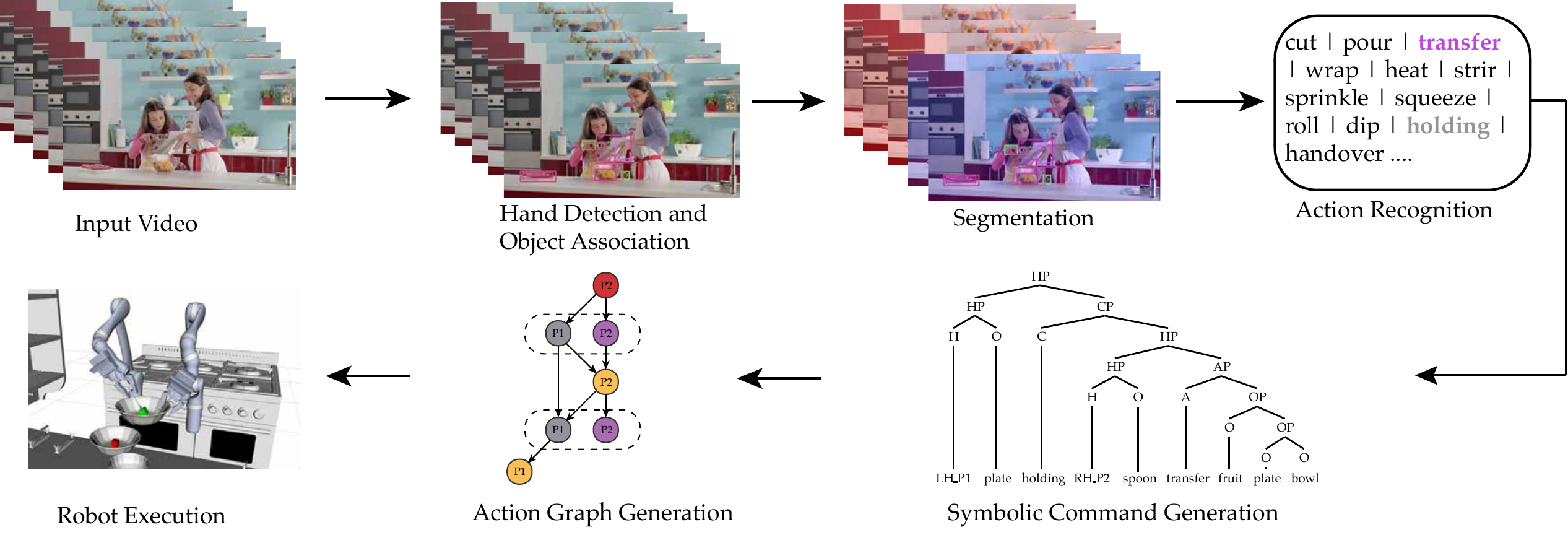}
\caption{Proposed system architecture.}
\label{fig:summary}
\vspace{-15pt}
\end{figure*}

We build upon this theory of language for action to propose a system for \textit{zero-shot} understanding both single and \textit{collaborative} actions from \textit{full-length} YouTube videos. Our key insight is that \textit{hands contain both spatial and temporal information of the demonstrated actions}. This allows using hand trajectories to temporally segment full-length videos to short clips, derive hand-object and object-object associations and infer the demonstrated actions.

Our system takes as input a YouTube video showing a collaborative task from start to end. We assume that the objects in the video are annotated with labels and bounding boxes, e.g., by running a YOLOR algorithm~\cite{wang2021you} (Fig.~\ref{fig:best}). We also assume a skill library that associates a detected action with skill-specific motion primitives. \emph{The YouTube video has not been seen before and no information from the video has been included as training set, thus we refer to our system as zero-shot imitating system.} We focus on cooking tasks because of the variety in manipulation actions and their importance in home service robotics. 

Fig.~\ref{fig:summary} shows the components of the proposed system. We make the following contributions: 
\begin{itemize}
\item We propose a context-free collaborative action grammar that generalizes previous work on action grammar~\cite{yang2015robot} to account for collaborative actions, e.g., a person holding a plate for someone else, and object-object associations, such as ``the chicken \textit{on} the chopping board.'' We use the grammar-based rules to derive action trees, that are more human-interpretable than deep neural network robot policies and reward functions generated by existing work on learning from videos~\cite{finn2017one,yu2018one,chen2021learning}, from the visual processing system described below.
\item We present a visual processing system for zero-shot action understanding from full-length videos. The system detects the human hands in the video and uses the hand trajectories to split the video into clips. It then associates objects with hands and objects with other objects spatially and temporally to identify which objects are manipulated. We use the commonsense knowledge extracted from large language corpus~\cite{marin2019learning,41880} to recognize individual actions. We refer to this action recognition module as a commonsense reasoning-based action recognition module. We also temporally track which person is manipulating each object, and whether the ownership of an object changes over time to identify and recognize collaborative actions. 
\item We propose an open-source platform for executing the generated action trees in both simulation and in the real world that concatenates the action trees generated by the context-free grammar parser.
\end{itemize}

We find that the commonsense reasoning-based action recognition module works well in the cooking domain. In the FOON dataset~\cite{paulius2018functional}, which has annotations prepared from 100 public domain YouTube cooking videos with a third-person view, the commonsense reasoning-based action recognition module achieved an average precision of $41\%$ and recall of $69\%$. We additionally annotate and show the performance of the whole system in six \textit{new, previously unseen} full-length YouTube videos that include collaborative actions between two persons. The precision and recall of the overall system was $0.58$ and $0.51$. We finally show a demonstration in simulation of two robots and in a proof-of-concept demonstration in the real world executing all the actions of one video using the open-source platform.

The current system is focused on cooking videos assigning to objects properties such as ``tools'', ``ingredients'' and ``containers.'' We leave investigating its applicability to other domains for future work. An additional limitation is that the extracted action sequences are executed in an open-loop manner and thus do not withstand real-world failures or disturbances.

Nevertheless, we are excited that this work brings us a step closer to having robots generate and execute a variety of semantically meaningful plans from watching cooking videos online.

\section{RELATED WORK}

In this paper we propose a system for converting full-length unconstrained videos from the Internet to collaborative action plans executed by one or more robots. Most relevant to ours is prior work on temporal activity segmentation and action understanding.

\noindent\textbf{Temporal Activity Segmentation.} Work in activity segmentation includes learning Gaussian Mixture Models~\cite{sener2018unsupervised}, specifying cost-functions incorporating spatial and temporal features of the trajectory~\cite{lasotabayesian} and combining classifier outputs with a symbolic grammar to parse sequence data~\cite{qi2018generalized}. Zhou et al. propose an end-to-end method for procedure segmentation~\cite{zhou2018towards}. More recently, temporal convolutional networks have been proposed to capture long-range dependencies for this task~\cite{wang2020boundary,li2021temporal,huang2020improving}. However these approaches rely on expensive data annotation. Thus more and more unsupervised and self-supervised learning approaches~\cite{wang2022sscap,li2021action} have been proposed to reduce the dependence on labeled data. 



While our system can accommodate any state-of-the-art temporal segmentation technique~\cite{li2021action,li2021temporal,wang2022sscap}, we applied the Greedy Gaussian segmentation algorithm by Hallac et al.~\cite{hallac2018greedy}. The algorithm is applicable to general multivariate time-series data and we adopt it for segmenting videos to clips using the extracted hand trajectories, based on the insight that hands contain temporal and spatial information about manipulation actions.


\noindent\textbf{Human Activity Understanding.} There has been a lot of work on human activity recognition~\cite{tang2020asynchronous,wu2019long,tirupattur2021modeling}. Recent work on deep learning approaches has enabled the generation of natural language~\cite{venugopalan2014translating}, individual robot commands~\cite{nguyen2018translating}, low-level robot policies~\cite{finn2017one,yu2018one,chen2021learning,DBLP:journals/corr/abs-1810-00146} and neural programs~\cite{sun2018neural} using manually annotated datasets or visual human demonstrations. Generation of collaborative actions has been achieved by representing them as social affordances~\cite{shu2017learning} from data recorded in a lab setting. Pastra et al.~\cite{pastra2012minimalist} discuss a minimalist grammar for action understanding, inspired by the suggestion by Chomsky~\cite{chomsky1993lectures}. Based on this theoretical insight, Yang et al.~\cite{yang2015robot,yang2014cognitive} proposed a system that uses deep neural networks for hand and object detection and association, while leverages a language corpus for action recognition. Performance was shown on 12 selected clips. In our preliminary work~\cite{hejia_isrr19}, we presented a collaborative grammar whose functionality was tested only qualitatively in a small number of manually selected simple clips, rather than in full-length unconstrained videos.

Contrary to the aforementioned work, our system includes a full pipeline that takes as input \textit{full-length}, previously unseen videos with annotated objects and bounding boxes, infers manipulated objects and tools based on hand detections, recognizes single and collaborative actions representing their structure using a \textit{collaborative} action grammar and concatenates the actions into a temporal sequence of executable commands. Importantly, the proposed system is modular so that individual components can be replaced or combined with state-of-the-art techniques, such as CNN-based methods for action recognition~\cite{Wang_2021_CVPR,Chen_2021_ICCV}. Moreover, different from previous work that learns to generate deep neural network robot policies~\cite{finn2017one,yu2018one} and reward functions~\cite{chen2021learning}, our framework generates human-interpretable action trees since human-interpretability is important for robots that work in human environments. 


\section{SYSTEM ARCHITECTURE}

The input to the system is a full-length, previously unseen video from the web. We assume that objects in the video are labeled and a bounding box is provided for each object e.g., using a state-of-the-art object detection algorithm~\cite{wang2021you}. We base this assumption on the tremendous progress of recent object-detection algorithms and the availability of large data-sets. This is the only labeled input data provided to the system. \footnote{While the current implementation of the system works for videos of one person working independently or two persons collaborating, an extension to three or more persons interacting in pairs is straightforward and left for future work.}

\noindent\textbf{Hand Detection.}
Our work is based on the insight that hands are the main driving force of manipulation actions~\cite{yang2014cognitive}. We use OpenPose~\cite{cao2018openpose}, which detects jointly the human body and hands. We use the detected hands to (1) segment videos by tracking the hand trajectory, and (2) detect which objects are manipulated at a given point in time.

\noindent\textbf{Video Segmentation.}\label{subsec:segmentation}
We temporally segment the video to short clips using the trajectories of the detected hands as time-series data, performing a separate segmentation for each hand of the actors in the video. The trajectory of each hand is a sequence of hand positions on the image frames. We use a greedy approach~\cite{hallac2018greedy}, which formulates the segmentation of a trajectory as a covariance-regularized maximum likelihood problem of finding the segment boundaries. The result segmentation for a hand is a list of sub-sequences of hand positions.  We then generate a new sequence of segments for the whole video as the union of individual segments for different hands, which we will use for action recognition. This method results in over-segmentation with some actions spanning multiple segments, which is common in segmentation algorithms~\cite{grundmann2010efficient}. Therefore, we merge segments with identical action trees in the action graph generation phase.




\noindent\textbf{Object Association.} 
After video segmentation, we extract objects that are relevant to actions in each segment. We do this by associating objects with hands and with other objects based on their relative positions in the frame. We introduce a semantic hierarchy of objects, by assigning them to three classes: \textit{tools} that manipulate other objects, e.g., knife and fork, \textit{containers} that can contain other objects, e.g., pot and bowl, and \textit{ingredients}, e.g., banana and lemon, that can not contain other objects. For robustness, we only keep the hand and object associations retained for at least a pre-defined number of consecutive frames.

\noindent\textit{Ingredient-Container Association.} We associate container objects with ingredients, if there is an overlap in the bounding boxes of the two. We use containers to detect transfer of objects from one container to the other, e.g. transfer a tomato from a bowl to a chopping board. We use the Jaccard index~\cite{jaccard1912distribution} between the bounding boxes of the two objects to pair an ingredient with a container that most likely contains that ingredient.


\noindent\textit{Hand-Object Association.} We want to detect the objects grasped by the hands and then propagate this association to nearby objects that can inform the action recognition. This allows us to infer which objects are directly manipulated or used as tools to manipulate other objects. We associate detected hands with objects whose bounding boxes overlap with the box of the hand. In the case of multiple overlaps, we associate the hand with the object whose bounding box is the closest to the bounding box of the hand. If the closest object is an ingredient that is contained by another object, we associate the hand with its container instead.

\noindent\textit{Object-Object Association.} 
For each object that has been associated with a hand, we look to associate that object with other objects that are possibly manipulated. For instance, if a hand grasps a spoon, we wish to see if the spoon is used to stir a pot nearby. We do this by associating objects associated with each hand with other containers or ingredients based on the Euclidean distance between the bounding boxes of the two objects. We don't associate the grasped object with any tool object based on the assumption that a tool object will not be the receiver of any action.

\begin{figure*}[t!]
\centering
\includegraphics[width=0.95\linewidth]{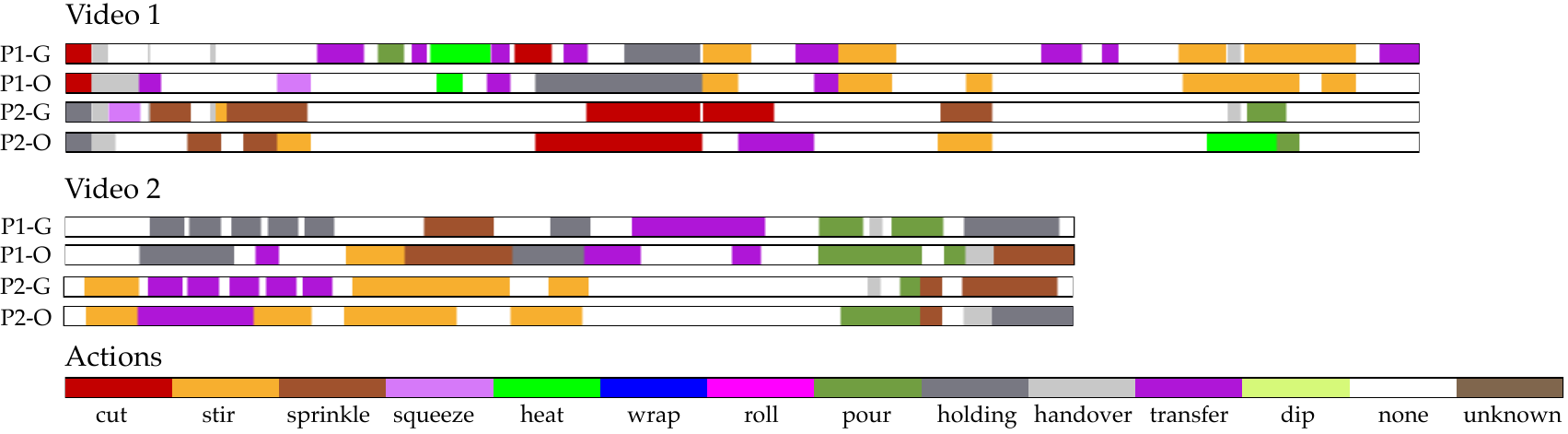}
\caption{Segmentation of the two test videos. For each video we show the ground truth for the adult (P1-G), our result for the adult (P1-O), the ground truth for the child (P2-G) and our result for the child (P2-O). The bottom color bar shows different colors for different actions.}
\label{fig:segmentation}
\vspace{-10pt}
\end{figure*}
 
\noindent\textbf{Action Recognition.}
After segmenting the video into clips and pairing objects with hands, we recognize actions performed by humans in the videos. We have two types of actions, actions performed by a single person, which we name \textit{individual or single actions}, and \textit{collaborative actions} performed by a pair of humans in the video. As a special case of individual actions, we introduce \textit{transfer} actions, which occur when an object moves from one container to another. This allows to detect transfer of an ingredient between containers.

\noindent\textit{Individual Actions.} We recognize commonsense actions~\cite{yang2015robot}, using a trained language model from a recipe corpus~\cite{marin2019learning}. Given a set of candidate actions and a set of candidate objects, we extract $P(Object | Action)$ for each possible bigram consisting of one object word and one action word in corpus. We then formulate action recognition task in terms of approximate sampling and prediction from the posterior distribution:


\setlength{\abovedisplayskip}{0pt}%
\setlength{\belowdisplayskip}{0pt}%
\setlength{\abovedisplayshortskip}{0pt}%
\setlength{\belowdisplayshortskip}{0pt}%
\setlength{\jot}{0pt}
\begin{align}
    P(A | O_1, ..., O_n) \sim \prod_{k=1}^n P(O_k | A)  P (A)
\end{align}


where $A$ is the performed action and $O_1, ... , O_n$ are the objects involved in the action respectively. We then select the most likely action. 




\noindent\textit{Transfer Actions.} We treat transfer actions separately from the other individual actions, since they occur when an object is moved from one container to the next and thus require tracking an object's association temporally. These actions are critical in keeping track of the location of the food in the cooking task.

\noindent\textit{Collaborative Actions.} We detect a collaboration: (1) when two persons grasp the same object, or (2) the object grasped by one person is used as a tool to manipulate an object grasped by another person. In case (1), we check over time which hands grasp the object and detect a \textit{handover} if the person grasping the object changes. Otherwise, we detect a \textit{holding} action, for instance when one person assists the other person stirring a pot by holding the pot as well.
{
\begin{figure}[t!]
  \centering
   \small
  \begin{align}
        \setcounter{equation}{0}
      &HP\ &\rightarrow    &\hspace{0.20in} H\ O\ |\ HP\ AP\ |\ HP\ CP\ |\ H\ OP\ |\ H\ AP \\
      &AP\ &\rightarrow &\hspace{0.20in} A\ O\ |\ A\ OP\ | \ A\ HP\\
      &CP\ &\rightarrow &\hspace{0.20in} C\ HP\\
      &OP\ &\rightarrow &\hspace{0.20in} O\ O\ |\ O\ OP \\
      &H\ &\rightarrow &\hspace{0.20in} Hand\\
      &C\ &\rightarrow &\hspace{0.20in} Collaboration\\
      &O\ &\rightarrow &\hspace{0.20in} Object\\
      &A\ &\rightarrow &\hspace{0.20in} Action
      \end{align}
  \caption{A Collaborative Manipulation Action Context-Free Grammar}
  \label{fig:intgrm}
  \vspace{-20pt}
\end{figure}
}

\noindent\textbf{Action Grammar Parsing.}
We need to represent the structure of the recognized actions for a robot to execute them. We use a manipulation action grammar~\cite{yang2015robot} which assumes that hands ($H$) are the driving force of both single manipulation actions ($A$) and collaborative actions ($C$). A hand phrase $HP$ contains an action phrase $AP$, or a collaborative action phrase $CP$. We also introduce an object phrase $OP$, which we use to indicate container - ingredient relationships between objects, e.g. a tomato in the bowl, as well as transfer actions from one container to another. The grammar is given in Fig.~\ref{fig:intgrm}. The rules (5)-(8) are terminal, with \textit{Hand} taking the values: ``LH\_P1'', ``RH\_P1'', ``LH\_P2'' and ``RH\_P2,'' ``LH\_P1'' being the left hand of the first person and so on. We use a context-free grammar parser~\cite{church1989stochastic} to parse the constructed visual sentences~\cite{hejia_isrr19} and output a parse tree of the specific manipulation action. The robot can then execute the action by reversely parsing the tree~\cite{yang2015robot}. Fig.~\ref{fig:trees} shows the constructed trees from five different action clips.




\begin{figure}[t!]
\centering
\includegraphics[width=0.35\linewidth]{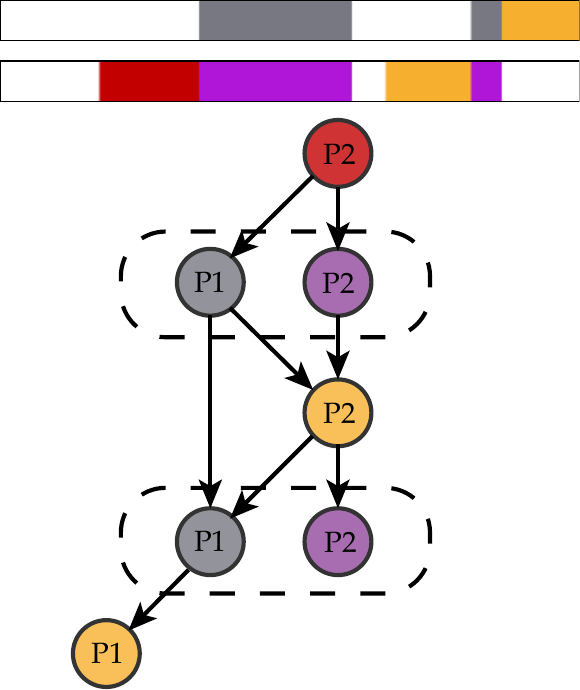}
\caption{Generated task graph for the (Rita) video. Different actions are shown in different colors using the colorbar of  Fig.~\ref{fig:segmentation}. $P_1$ indicates an action executed by the woman in the video and $P_2$ by the girl. The dotted line indicates a collaboration between the two actors.}
\label{fig:demo}
\vspace{-20pt}
\end{figure}

\noindent\textbf{Action Graph Generation and Execution.}
Because of over-segmentation, we end up with multiple consecutive segments that are parts of the same action. Therefore, we first merge consecutive segments from the video with identical actions and manipulated objects. We do not require identical ingredient objects, since they may not be visible in some of the segments.

We then generate an action graph that combines the generated action trees to action sequences, each corresponding to each person in the video. We then decompose each action into motion primitives. We define four primitives~\cite{holladayforce}: grasp, engage, actuate and place. For instance, a transfer action of a food from a plate to a bowl with a spoon includes grasping the spoon (grasp), moving it close to the food (engage), performing the scooping motion (actuate), moving the spoon close to the bowl (engage), turning it to remove the food (actuate), and placing it back in its initial position (place). We use Task Space Regions (TSRs)~\cite{Berenson:2011:TSR:2046796.2046797} to specify feasible regions of target poses of the robot's end effector in the grasp, engage and place primitives and use bidirectional rapidly-exploring random trees~\cite{lavalle2001rapidly} to plan collision-free paths.

We implement the action graph as an open-source platform, that enables collaborative task execution in the cooking domain.\footnote{\url{https://github.com/icaros-usc/wecook}} Fig.~\ref{fig:demo} shows the action graph generated for the (Rita) video.\footnote{Rita:~\url{https://www.youtube.com/watch?v=d3SZH7NFDjc&t=69s}} The graph includes one action (stir) that was not detected by the visual processing system.

\section{EXPERIMENTS}

Our first experiment evaluates the commonsense reasoning-based module for recognizing single actions in the cooking domain based on a large annotated dataset of 100 YouTube videos that include single-person actions. Equipped with these results, we test our system and compare it with a baseline system that consists of a state-of-the-art action detection framework~\cite{tang2020asynchronous} on six YouTube videos that show two persons collaborating in the various cooking tasks.

\subsection{Action Recognition with Commonsense Reasoning} Based on the fact that in cooking people use specific tools for specific actions, such as a knife to cut and a rolling pin to roll, we hypothesize that the system will be able to recognize most actions from the objects and tools in the scene.

\noindent\textbf{Dataset and Experiment Settings.} 
We use the publicly available manipulation knowledge representation dataset called the Functional Object-Oriented Network (FOON)~\cite{paulius2018functional}. The dataset was prepared from 100 public domain YouTube cooking videos with a third-person view, and it includes annotated actions and objects involved in these actions. We restricted the action set to a set of common candidate actions similarly to previous work~\cite{yang2015robot}. We trained our language model on the Recipe1M+ dataset~\cite{marin2019learning} and the One-Billion-Words dataset~\cite{41880}. 
\begin{figure}[t!]
  \centering
   \includegraphics[width=0.95\linewidth]{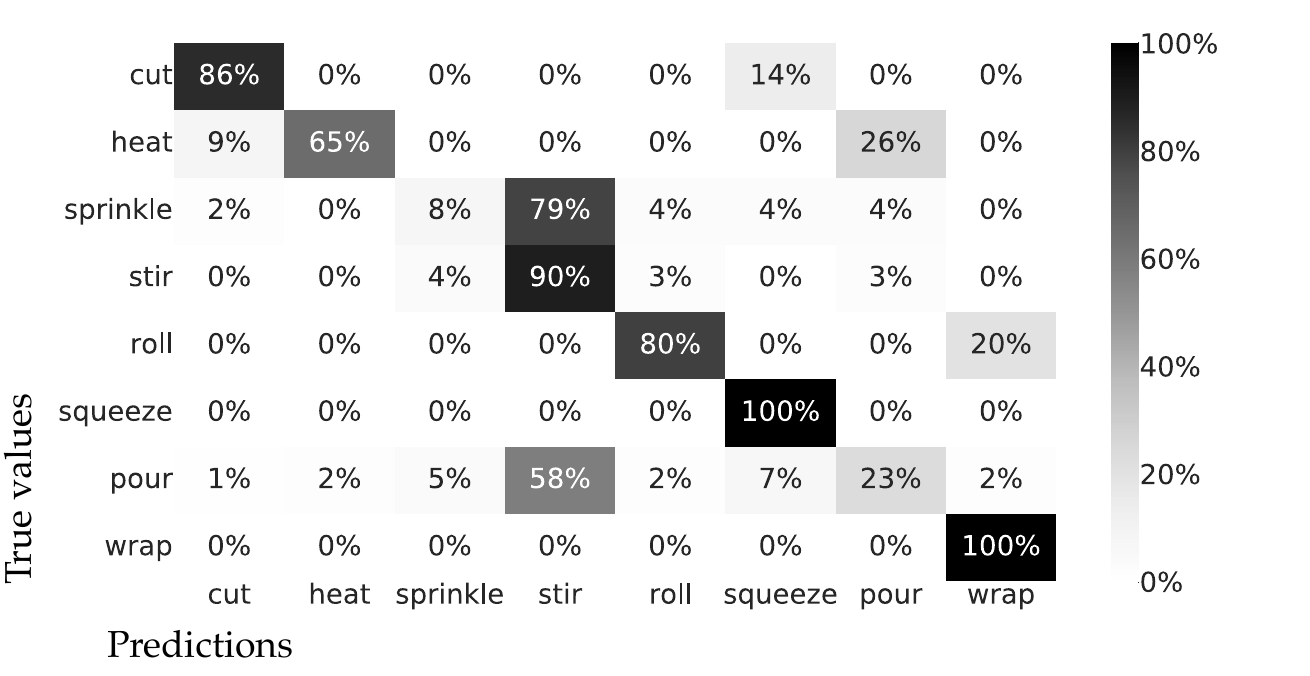}
  \caption{Normalized confusion matrix of the action recognition with commonsense reasoning. The \textit{rows} indicate \textit{true values}, while the \textit{columns} are the \textit{predictions} of the action recognition module.}
  \label{fig:confusion_matrix}
\vspace{-20pt}
\end{figure}

\noindent\textbf{Results.} The average precision over all actions was $41\%$ and the recall was $69\%$. Fig.~\ref{fig:confusion_matrix} shows the normalized confusion matrix for our action set, with the true positive rate (recall) for each action. The action \textit{sprinkle} and action \textit{pour} were often misclassified as \textit{stir}, because of the prevalence of container objects, such as bowl and pot, which skewed the inference towards the stir action. On the other hand, the performance for \textit{cut} was high because of the presence of the \textit{knife} tool. These results show the effectiveness of the commonsense reasoning-based module in the cooking domain.



\newcommand{\specialcell}[2][c]{%
  \begin{tabular}[#1]{@{}c@{}}#2\end{tabular}}

\begin{table}
\centering
\newcolumntype{C}[1]{>{\centering\let\newline\\\arraybackslash\hspace{0pt}}m{#1}}
\newcolumntype{R}{>{\raggedleft\arraybackslash}p{2cm}}
\newcolumntype{L}{>{\raggedright\arraybackslash}p{1cm}}
\caption{Comparison of the proposed system with a baseline system regarding the precision, recall and number of correctly detected actions in the collaborative cooking videos.}
\label{table:table-clearing-payoff}
\vspace*{-6pt}
\resizebox{0.7\columnwidth}{!}{
\begin{tabular}{lllllll}
 \hline
 

     &\multicolumn{2}{c}{Precision} & \multicolumn{2}{c}{Recall} & \multicolumn{2}{c}{Correct}    \vspace{0.5em}
   \\ 
   \cmidrule(lr){2-3} \cmidrule(lr){4-5} \cmidrule(lr){6-7}
& Ours & Ap1 & Ours & Ap1 & Ours & Ap1 \\
   \hline
  \specialcell{Sal} & $\textbf{0.52}$ & $0.31$ & $\textbf{0.48}$ & $0.35$ & $\textbf{11}$ & $8$\\ 
  \specialcell{Robalo} & $\textbf{0.38}$ & $0.22$ & $\textbf{0.50}$ & $0.22$ & $\textbf{9}$ & $4$\\ 
  \specialcell{Rita} & $\textbf{1.}$ & $0.30$ & $\textbf{0.83}$ & $0.50$ & $\textbf{5}$ & $3$ \\ 
  \specialcell{Peixe} & $\textbf{0.65}$ & $0.18$ & $\textbf{0.76}$ & $0.18$ & $\textbf{13}$ & $3$\\ 
  \specialcell{Massa} & $\textbf{0.62}$ & $0.60$ & $\textbf{0.41}$ & $0.19$ & $\textbf{13}$ & $6$ \\ 
  \specialcell{Rissois} & $\textbf{0.68}$ & $0.33$ & $\textbf{0.43}$ & $0.23$ & $\textbf{13}$ & $7$\\ 
  \specialcell{Total} & $\textbf{0.58}$ & $0.30$ & $\textbf{0.51}$ & $0.25$ & $\textbf{64}$ & $31$ \\ \hline
 \end{tabular}
 }
\label{tab:results}
 \vspace{-0.2cm}
\end{table}

\begin{table}
\centering
\caption{Number of Correct Trees for Different Types of Error. \\  A: Action Recognition Error, HO: Hand-Object Association Error, OO: Object-Object Association Error}
\label{table:table-clearing-payoff}
\vspace*{-6pt}
 \resizebox{0.65\columnwidth}{!}{
 \begin{tabular}{l r r r r r}
 \hline
 
 \hline
    & Total & Correct & A & HO   & OO  \\
    \hline
   \specialcell{Sal} & $21$ & $11$ & $8$ & $1$ & $1$ \\
   \specialcell{Robalo} & $24$ & $9$ & $5$ & $7$ & $3$ \\
   \specialcell{Rita} & $5$ & $5$ & $0$ & $0$ & $0$ \\
   \specialcell{Peixe} & $20$ & $13$ & $0$ & $3$ & $4$ \\
   \specialcell{Massa} & $21$ & $13$ & $4$ & $2$ & $2$ \\
   \specialcell{Rissois} & $19$ & $13$ & $3$ & $1$ & $2$ \\
   \specialcell{Total} & $110$ & $64$ & $20$ & $14$ & $12$ \\ 
 \end{tabular}
 }
\label{tab:errors}
 \vspace{-20pt}
\end{table}

\subsection{Collaborative YouTube Videos}
The FOON dataset does not include object bounding boxes and does not have videos with collaborative actions. We show the applicability of the entire system, shown in Fig.~\ref{fig:summary}, in six public YouTube videos, where we manually did the object annotation.\footnote{Masa:~\url{https://www.youtube.com/watch?v=1p2wBBmhPmk},\\
Rissois:~\url{https://www.youtube.com/watch?v=jAhQfH1PspU},\\
Sal:~\url{https://www.youtube.com/watch?v=GYr9Mw4ml5U},\\
Robalo:~\url{https://www.youtube.com/watch?v=wRBgiq_kzsY},\\
Peixe:~\url{https://www.youtube.com/watch?v=Jm77G7Xycyw},\\
Rita:~\url{https://www.youtube.com/watch?v=d3SZH7NFDjc}} Although there is no existing complete system that learns the collaborative manipulation plans like ours, we construct a baseline system that consists of a state-of-the-art action detection framework~\cite{tang2020asynchronous} and compare our system with it. The results show that our system achieves significantly higher action detection accuracy.


\begin{figure*}[t!]
\centering

\includegraphics[width=0.9\linewidth]{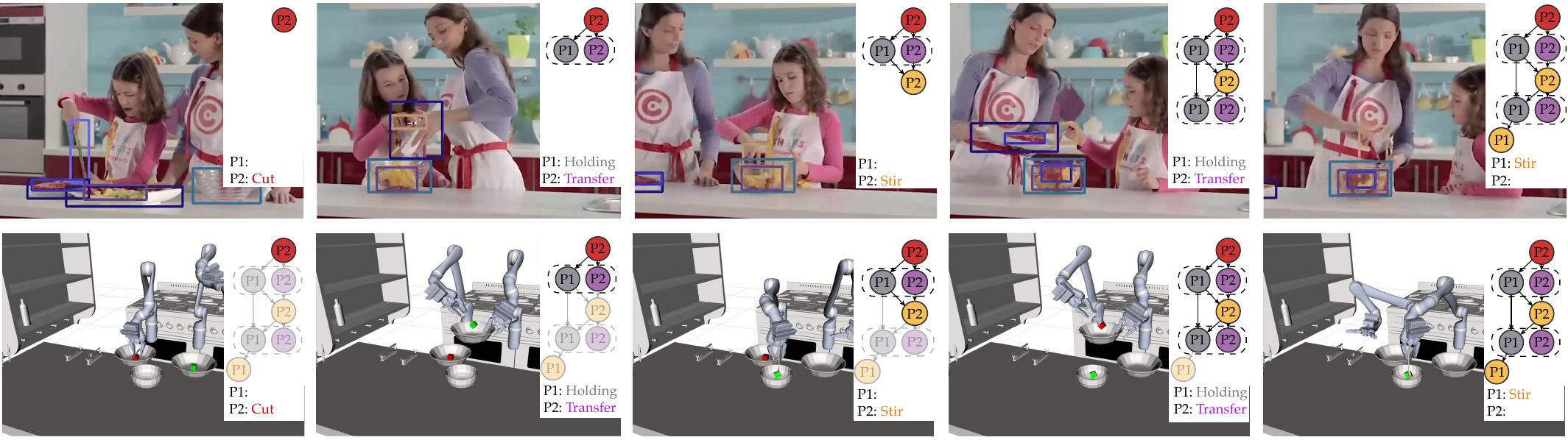}

\caption{Example frames of the video (Rita) showing two humans cooking together and corresponding snapshots in simulation environment of two robots executing the same actions. We visualize the learned and executed action graphs on the upper right corner of each frame and snapshot.}
\label{fig:video}
\vspace{-15pt}
\end{figure*}
\begin{figure}[t!]
\centering
\begin{subfigure}[b]{\linewidth}
\begin{tabular}{cc}
\raisebox{1ex-\height}{\includegraphics[width=.29\linewidth]{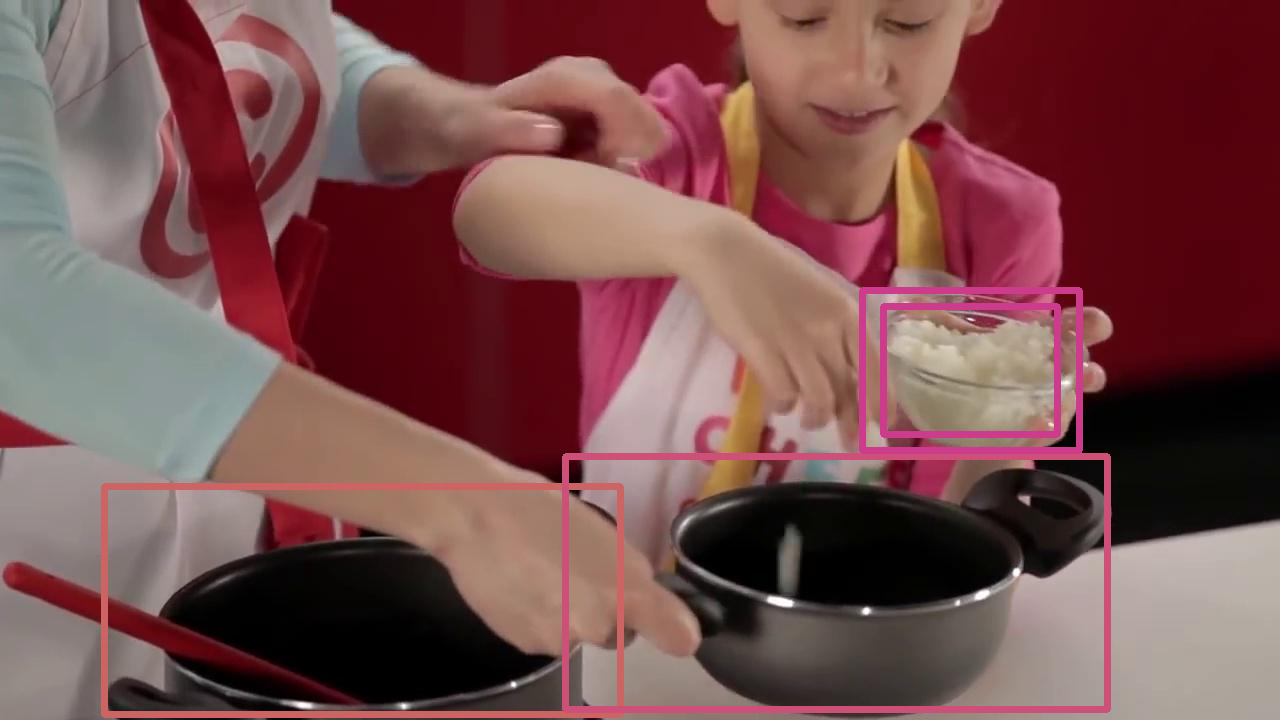}}\null\hfill&
\raisebox{-6ex-\height}{\small No tree generated}
\end{tabular}
\caption{The girl is sprinkling onion to the pot.}
\label{fig:no-obj}
\end{subfigure}
\begin{subfigure}[b]{\linewidth}
\begin{tabular}{cc}
\raisebox{1ex-\height}{\includegraphics[width=.29\linewidth]{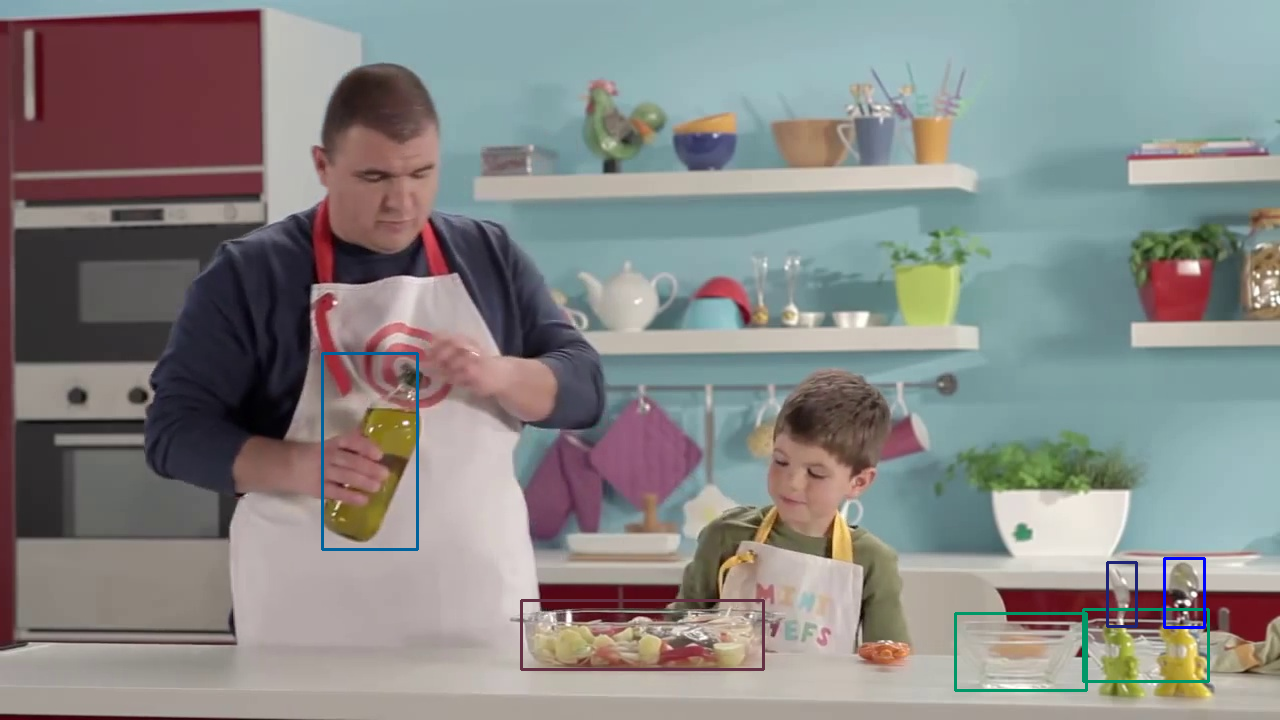}}\null\hfill&
\raisebox{1ex-\height}{\includegraphics[width=.33\linewidth]{obj-obj.tikz}}
\end{tabular}
\caption{The man is placing the olive oil back.}
\label{fig:obj-obj}
\end{subfigure}
\begin{subfigure}[b]{\linewidth}
\begin{tabular}{cc}
\raisebox{1ex-\height}{\includegraphics[width=.29\linewidth]{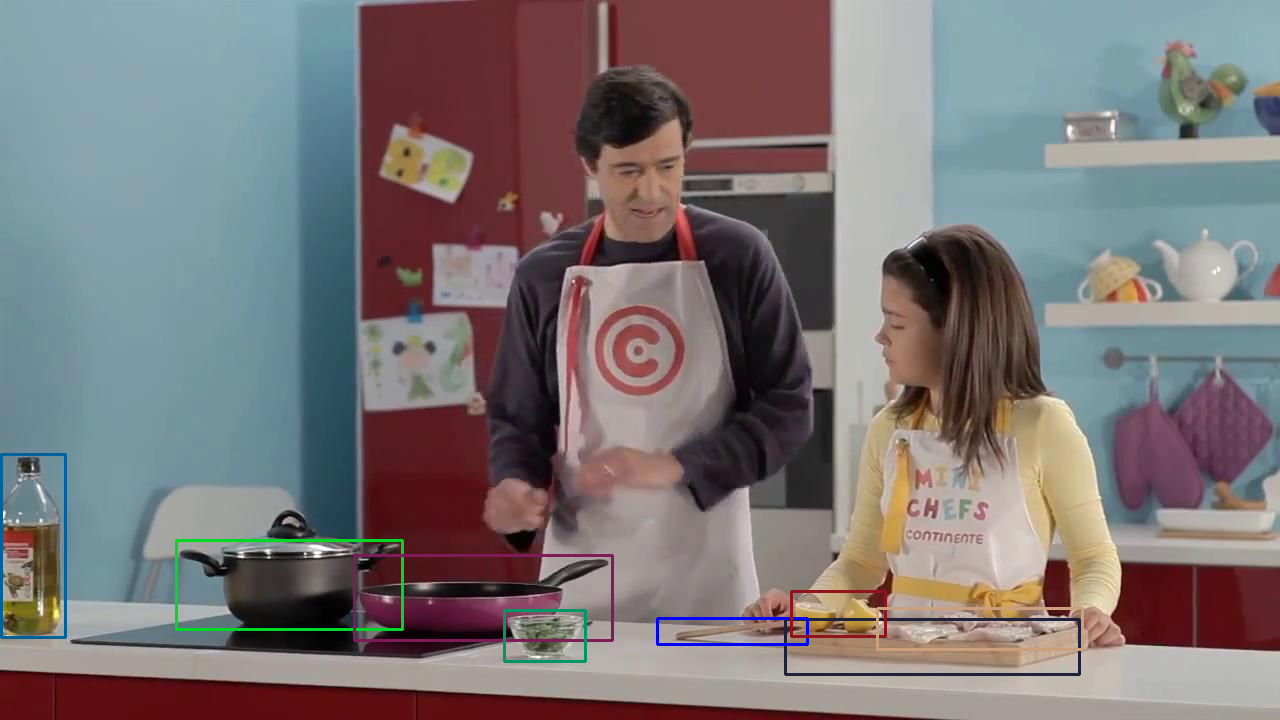}}\null\hfill&
\raisebox{1ex-\height}{\includegraphics[width=.33\linewidth]{hand-obj.tikz}}
\end{tabular}
\caption{The girl is not performing any action.}
\label{fig:hand-obj}
\end{subfigure}

\caption{Example frames and generated trees of 3 failure cases. The captions depict the groundtruth descriptions of each case.}
\label{fig:failures}
\vspace{-20pt}
\end{figure}

\noindent\textbf{Dataset and Experiment Settings.}
We set up a start and end time for each video, annotated the objects and set bounding boxes. We only annotated objects that were clearly visible, skipping objects that were heavily occluded. For training the baseline system, we annotated the action labels for each one second clip of the videos and human body bounding boxes for each frame, similar to the AVA dataset~\cite{gu2018ava} used in the baseline method paper~\cite{tang2020asynchronous}. The videos include a total of $37030$ frames and $126$ executed actions of $12$ different action types.

\noindent\textbf{Baseline System (Ap1).} We construct a baseline system based on AIA~\cite{tang2020asynchronous}, a state-of-the-art end-to-end deep learning-based action detection framework. We select AIA as our baseline method because it focuses on reasoning over person-person, person-object interactions. Given the ground truth object and human body annotations, the baseline system detects actions of each person in each one second clip of the test videos using AIA. The system then aggregates the results for each video segment. The video segments are generated using the video segmentation module of the proposed system for a fair comparison. We use the pre-trained model, which is pre-trained on Kinetics-700 dataset~\cite{carreira2017quo} for action classification task, proposed in the AIA paper~\cite{tang2020asynchronous}. For each test video, we then fine-tune the pre-trained model on the other $5$ videos. \footnote{Dataset and code for training and testing the baseline system is available at \url{https://bit.ly/3v524w6}.}



\noindent\textbf{Results.} We evaluate the performance of the system with respect to the percentage of correctly learned action trees. We define a correct action tree when the structure and the included persons, actions and objects are identical to the ground-truth~\cite{yang2015robot}, and the segment corresponding to that tree has a non-zero temporal overlap with the ground-truth segment. We specify the \textit{precision} as the number of action trees the system returns correctly out of the total number of detected instances, and the \textit{recall} as the number of action trees the system returns correctly out of the total number of ground-truth trees.

\begin{figure*}[t!]
\centering
\begin{subfigure}[b]{\linewidth}
\centering
\begin{tabular}{ccc}
\raisebox{1ex-\height}{\includegraphics[width=.23\linewidth,height=2.cm]{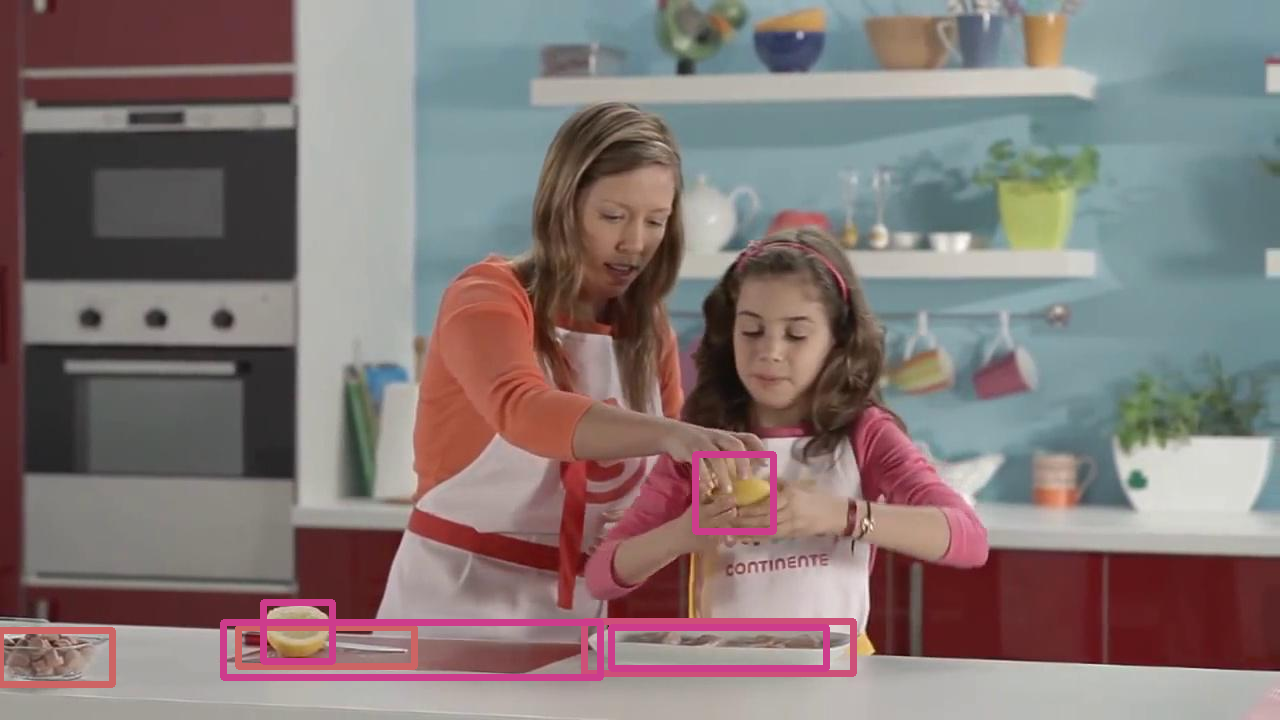}}\null\hfill&
\raisebox{1ex-\height}{\includegraphics[width=.23\linewidth,height=2.cm]{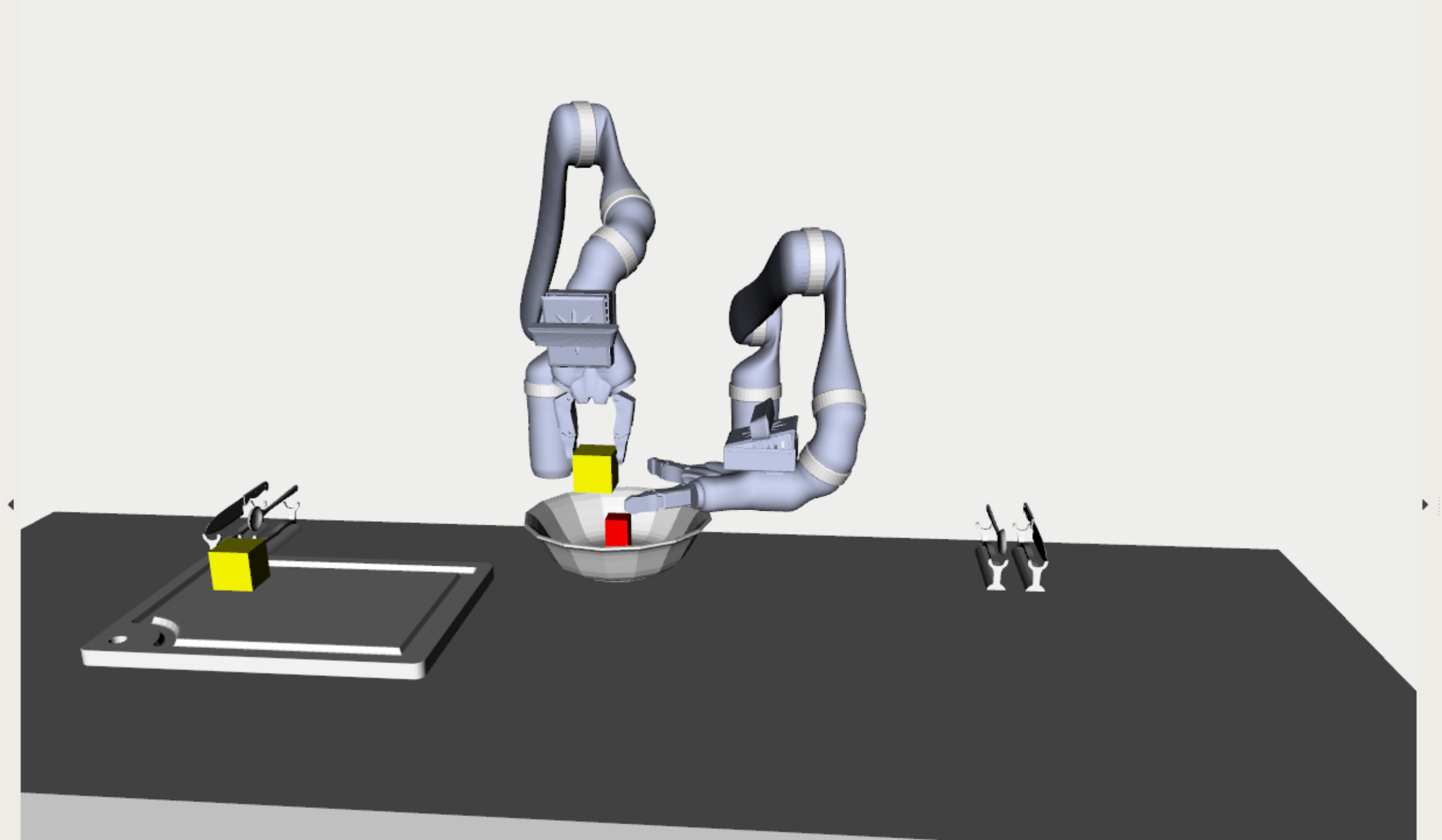}}\null\hfill&
\raisebox{1ex-\height}{\includegraphics[width=.25\linewidth]{handover-lemon.tikz}}
\end{tabular}
\caption{The woman is handing over a lemon to the girl.}
\end{subfigure}
\begin{subfigure}[b]{\linewidth}
\centering
\begin{tabular}{ccc}
\raisebox{1ex-\height}{\includegraphics[width=.23\linewidth,height=2.cm]{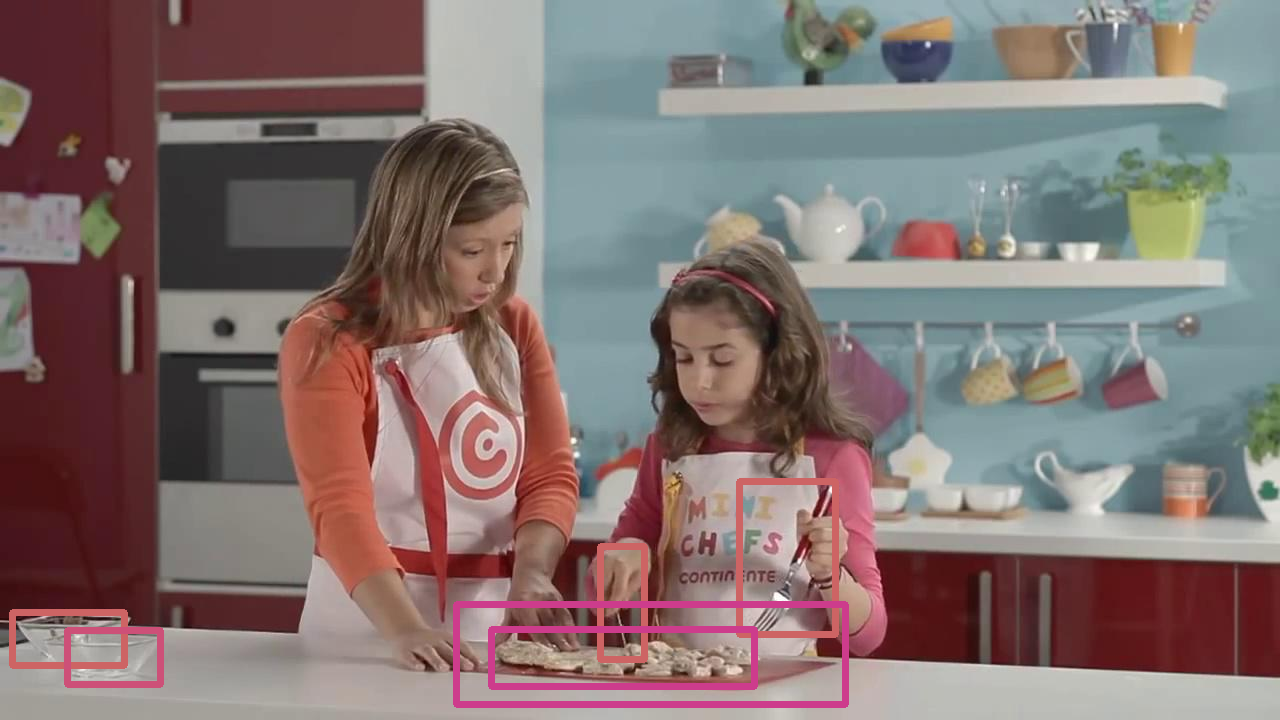}}\null\hfill&
\raisebox{1ex-\height}{\includegraphics[width=.23\linewidth,height=2.cm]{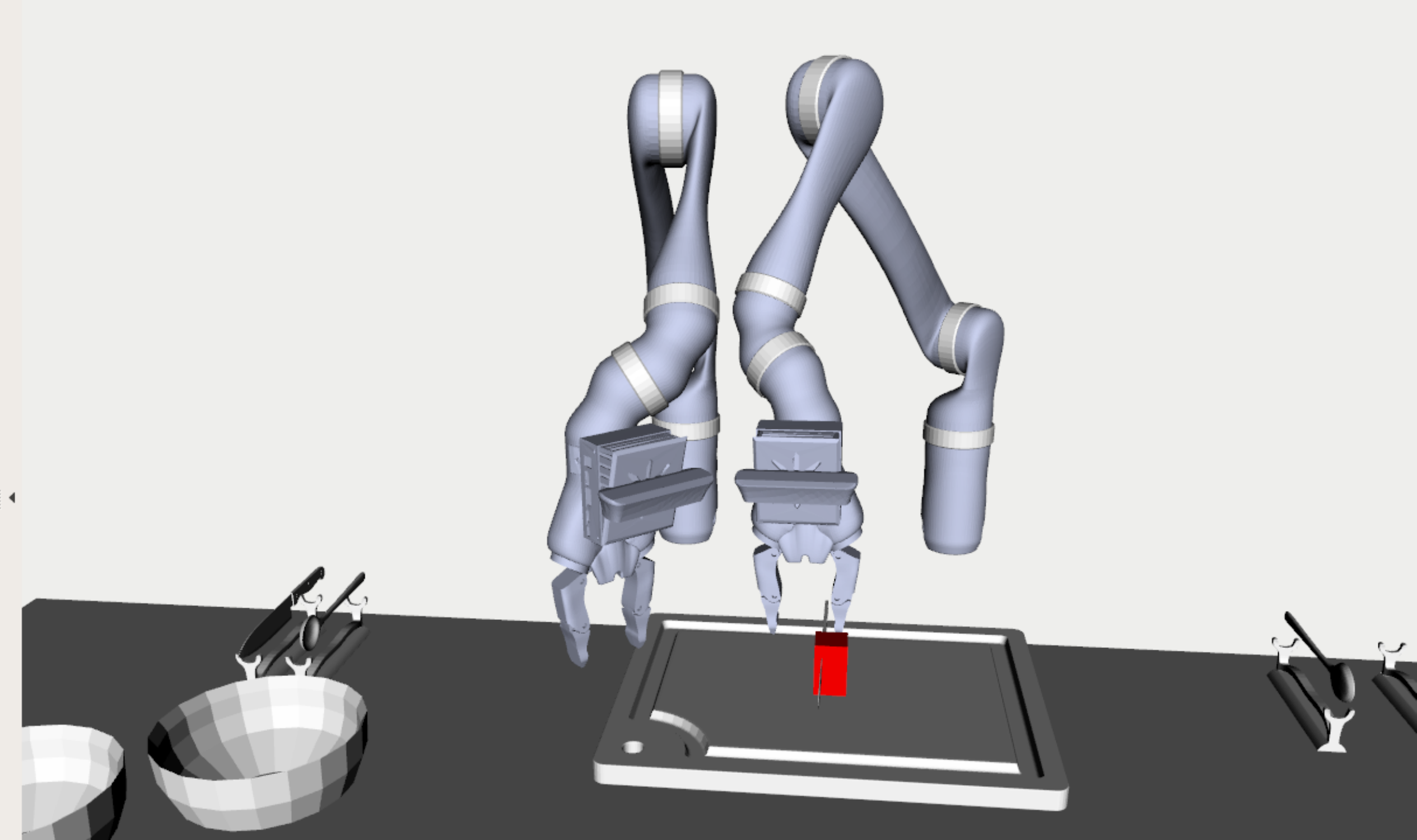}}\null\hfill&
\raisebox{1ex-\height}{\includegraphics[width=.25\linewidth]{holding-cut.tikz}}
\end{tabular}
\caption{The woman is holding the chopping board for the girl to cut the meat.}
\end{subfigure}
\begin{subfigure}[b]{\linewidth}
\centering
\begin{tabular}{ccc}
\raisebox{1ex-\height}{\includegraphics[width=.23\linewidth,height=2.cm]{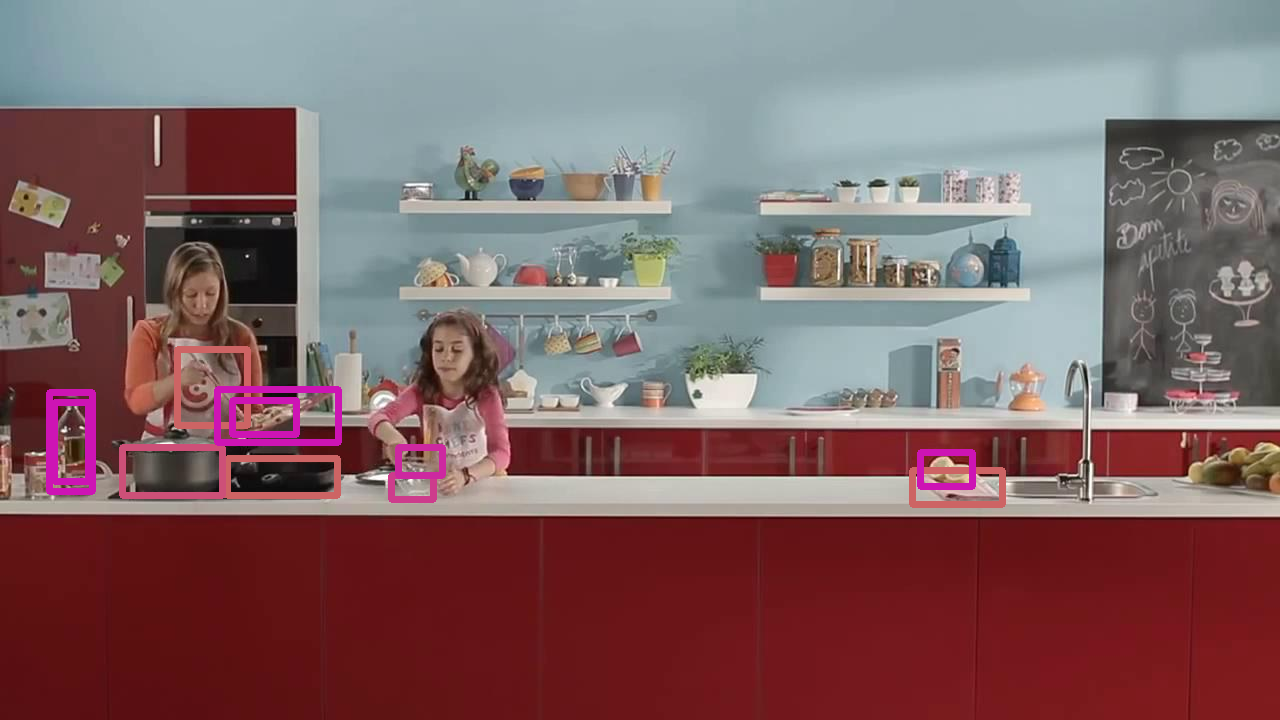}}\null\hfill&
\raisebox{1ex-\height}{\includegraphics[width=.23\linewidth,height=2.cm]{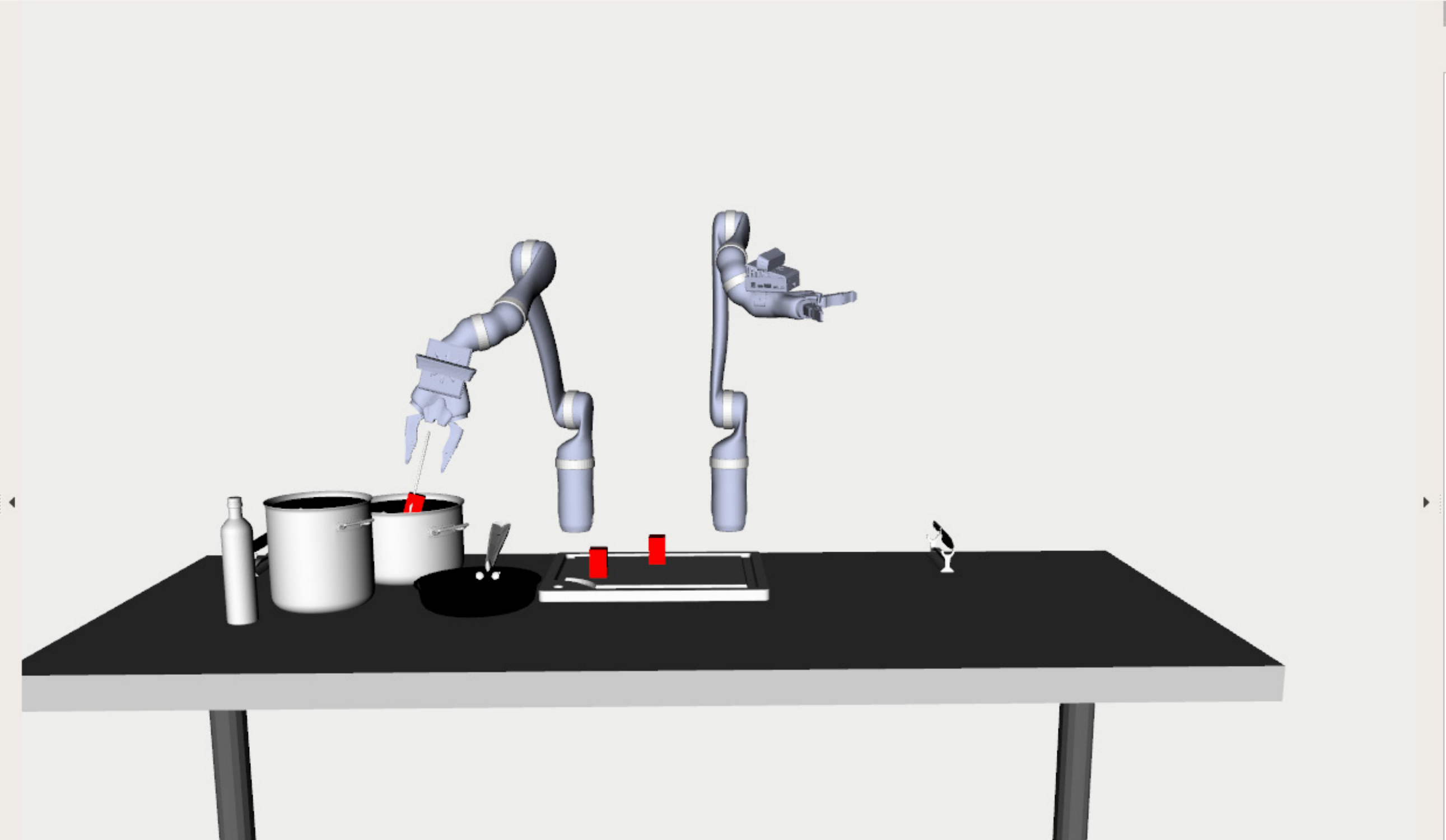}}\null\hfill&
\raisebox{1ex-\height}{\includegraphics[width=.25\linewidth]{transfer-only.tikz}}
\end{tabular}
\caption{The woman is transferring meat from the chopping board to the pot.}
\end{subfigure}
\begin{subfigure}[b]{\linewidth}
\centering
\begin{tabular}{ccc}
\hspace{25pt}\raisebox{1ex-\height}{\includegraphics[width=.23\linewidth,height=2.cm]{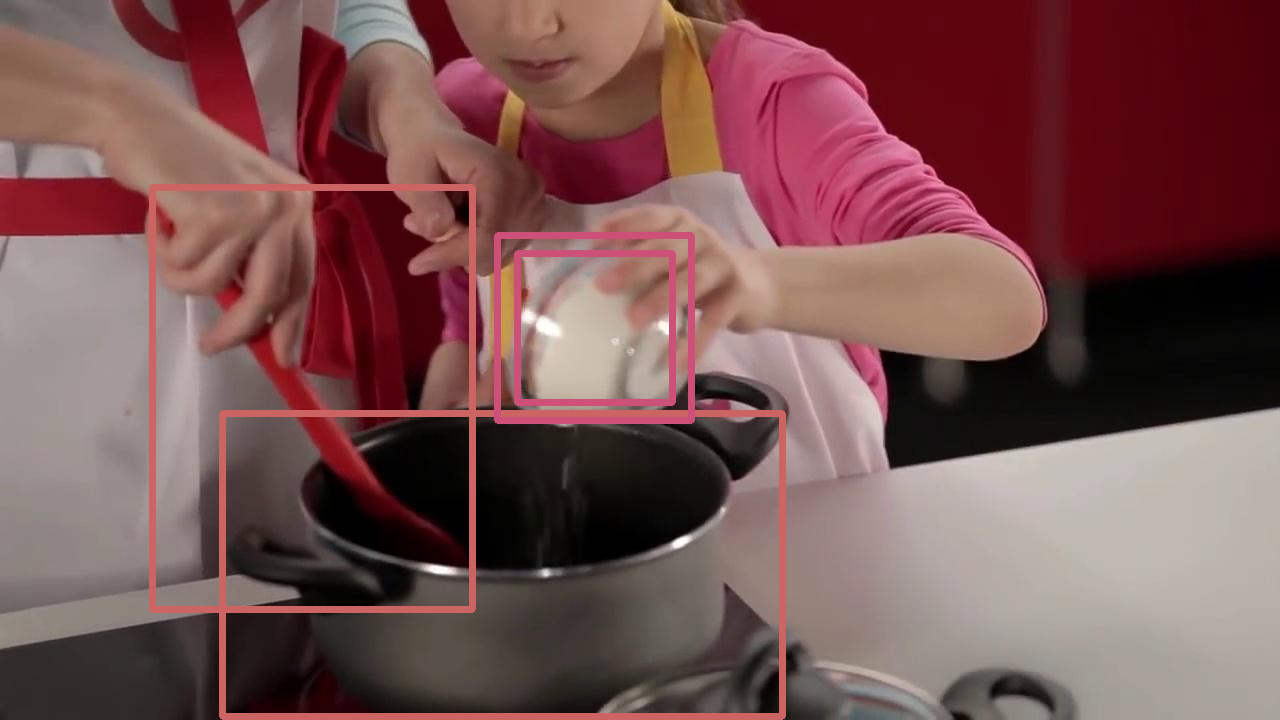}}\null\hfill&
\raisebox{1ex-\height}{\includegraphics[width=.23\linewidth,height=2.cm]{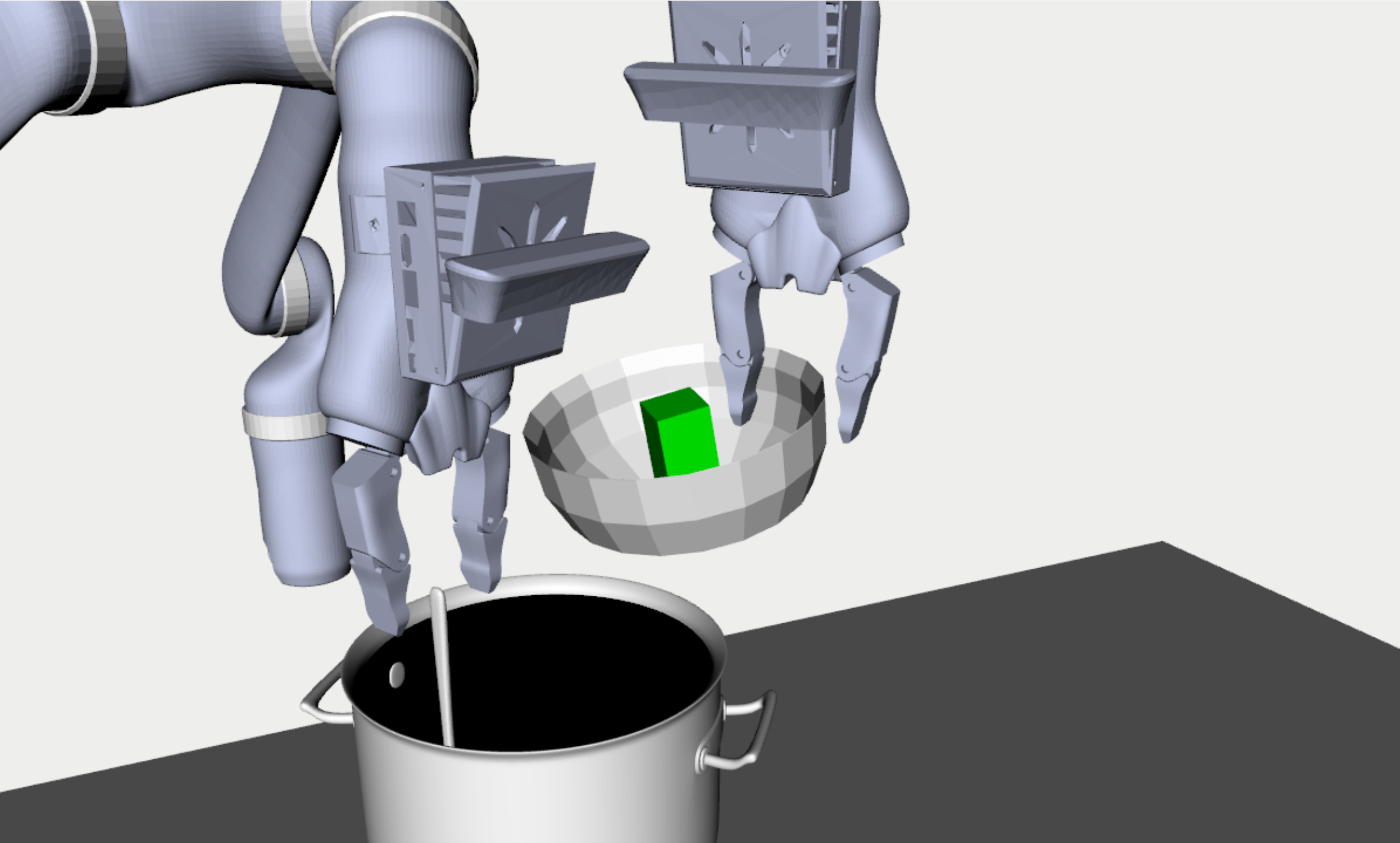}}\null\hfill&
\raisebox{1ex-\height}{\includegraphics[width=.15\linewidth]{close-up_1.tikz}}
\raisebox{1ex-\height}{\includegraphics[width=.2\linewidth]{close-up_2.tikz}}\hspace{-25pt}
\end{tabular}
\caption{The woman is stirring the pot while the girl is transferring the flour from the bowl to the pot.}
\end{subfigure}
\begin{subfigure}[b]{\linewidth}
\centering
\begin{tabular}{ccc}
\raisebox{1ex-\height}{\includegraphics[width=.23\linewidth,height=2.cm]{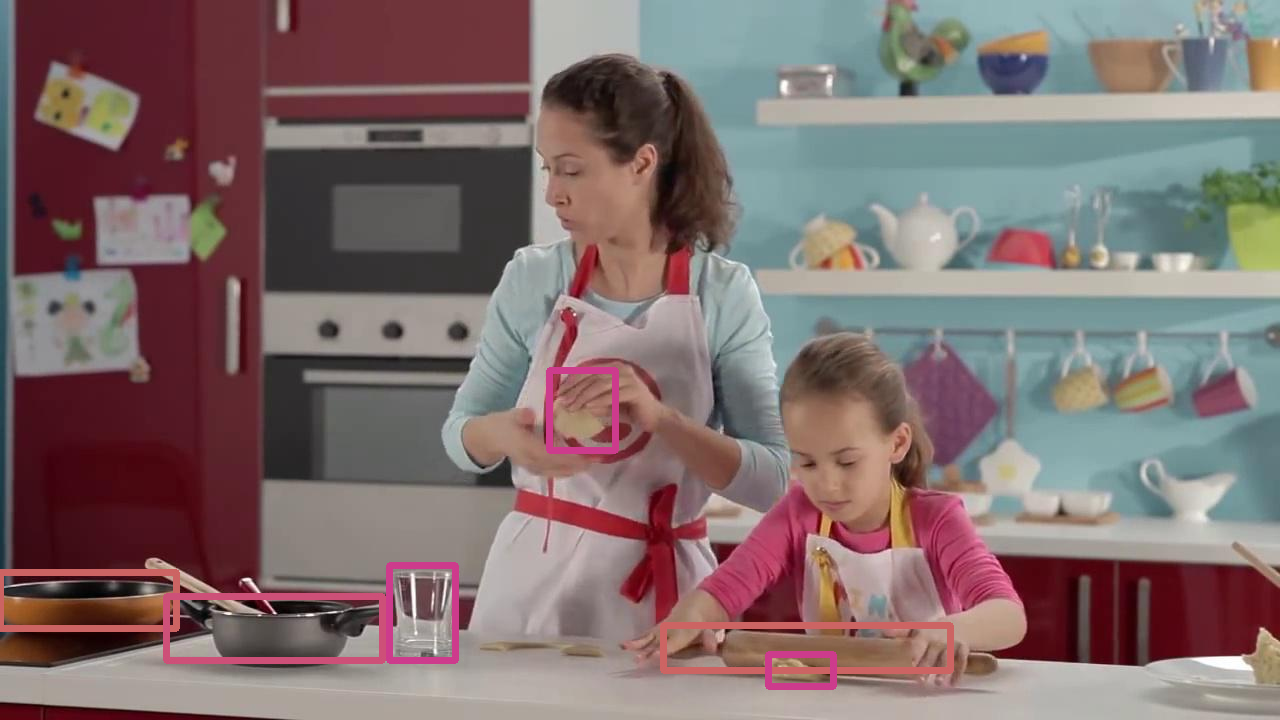}}\null\hfill&
\raisebox{1ex-\height}{\includegraphics[width=.23\linewidth,height=2.cm]{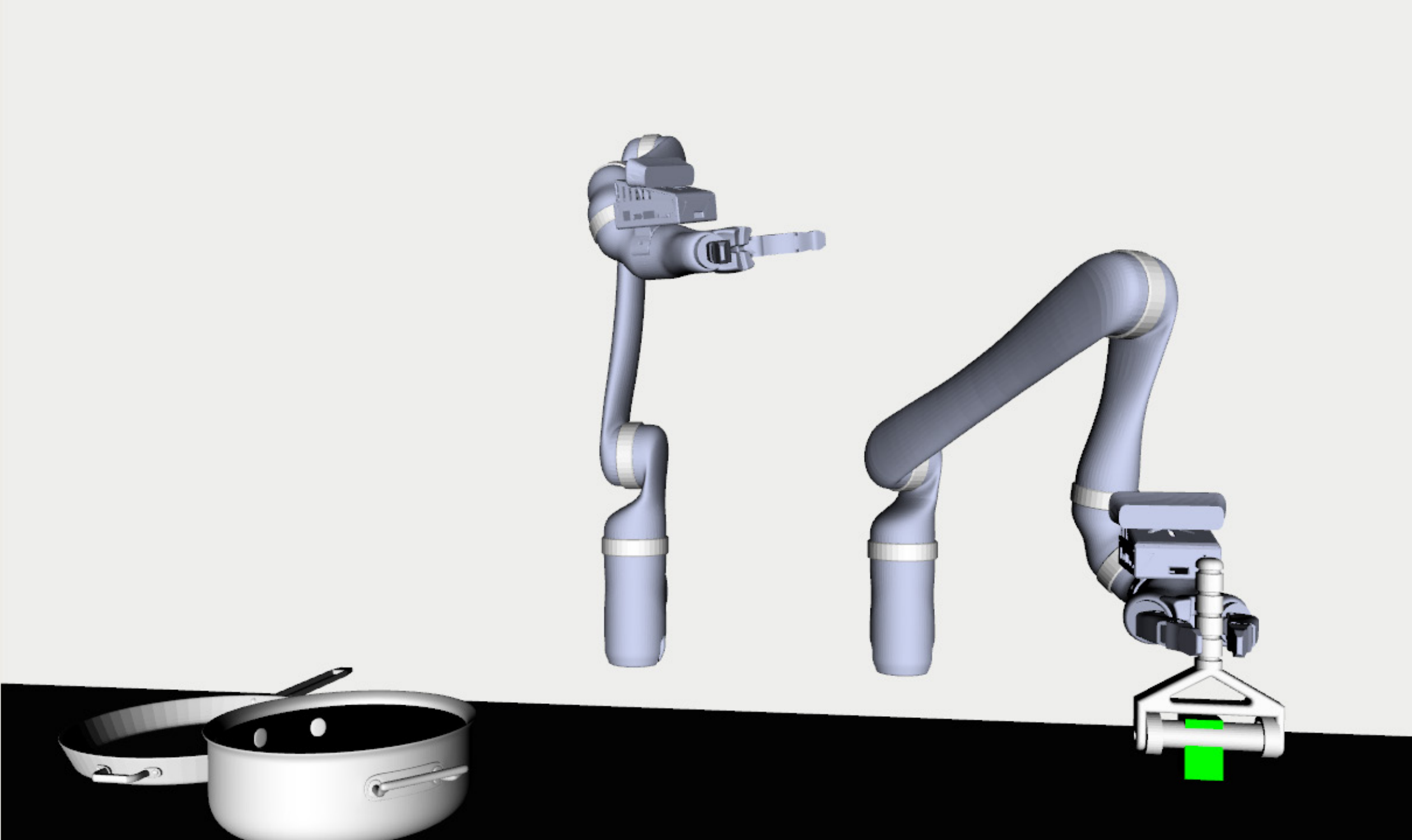}}\null\hfill&
\hspace{10pt}\raisebox{0ex-\height-2ex}{\includegraphics[width=.2\linewidth]{roll.tikz}}\hspace{20pt}
\end{tabular}
\caption{The girl is rolling the dough.}
\end{subfigure}
\caption{Example frames, snapshots in simulation environment of two robots executing the same actions with humans and generated actions trees of $5$ successful cases. The captions depict the ground-truth descriptions of each successful case.}
\label{fig:trees}
\vspace{-15pt}
\end{figure*}

Fig.~\ref{fig:trees} shows the action-trees and snapshots of the action tree executions by two robotic arms in the open-source platform. The proposed method reproduces successfully a variety of individual and collaborative actions.



Table~\ref{tab:results} shows that the proposed system achieved $0.58$ precision and $0.51$ recall while the baseline system (Ap1) only achieved $0.30$ precision and $0.25$ recall. It should be noted that for Ap1 a detection counts as correct if the output action name is correct since Ap1 only generates action names instead of executable action trees. Our system works better mainly because 1) our action detection module leverages commonsense knowledge to recognize actions while Ap1 detects actions only based on the visual features of the objects and human bodies thus it requires huge amount of annotated data, which is expensive to obtain, to fine-tune the pre-trained model; 2) By tracking hand-object and object-object associations, our approach is better at detecting collaborative actions and transfer action.

\noindent\textbf{Post-hoc Analysis.} We perform a post-hoc analysis of the data to identify different causes of failures of our system in the generated trees. Table~\ref{tab:errors} shows the number of correct trees that we generated and the number of trees that were erroneous for different reasons. We identified three sources of error: 1) action recognition (A) 2) hand-object associations (HO): some objects were not visible and we did not include them in the object annotations. There were also cases where the hand would be close to an object without grasping it, resulting in error in hand-object association. 3) object-object associations (OO): similarly to the hand-object association, the bounding boxes of two objects would overlap for consecutive frames, even though the objects were not interacting.

We observe that a main source of error was hand-object associations. This occurred often when an object was not visible and we did not include it in the object annotations. For instance, in Fig.~\ref{fig:no-obj}, the system failed to generate an action tree for the sprinkle action since the chopped onion was not annotated. A second source of error in hand-object associations was because of the distance between bounding boxes. For example, in Fig.~\ref{fig:hand-obj}, the system incorrectly infers that the girl is grasping the spoon, although the girl's hand does not make actual contact with the spoon. Errors occurred also in the object-object associations; for example, in Fig.~\ref{fig:obj-obj}, the system incorrectly associates the oil with the dish, although the man is bringing the object back.

\section{DEMONSTRATION}
To demonstrate the applicability of our system, we selected the video (Rita) where our system achieved perfect precision (Table~\ref{tab:results}), while it only missed recognizing one action. Fig.~\ref{fig:video} shows the execution of the generated action trees from that video by two simulated Kinova Gen2 lightweight robotic arms in the open source platform. In the accompanying video, we show also a proof-of-concept execution of the same actions with a robot and a human, using our open-source software platform.

\section{DISCUSSION}

\noindent\textbf{Limitations.}
 Our work is limited in many ways.
 As we observe in the post-hoc analysis section, our assumptions in how objects are associated with hands and with other objects could be limiting in the case of occlusions during collaborative actions or clutter. To deal with the occlusion, we could incorporate tracking modules for objects as we are using tracking modules for human hands. Detecting grasping types could improve the inferred associations~\cite{yang2015grasp}. Additionally, the YouTube videos were meant to be instructional and thus relatively clear. While our hand and object detection-based action prediction can be robust against implicit or ambiguous actions, language corpus-based commonsense reasoning will fail in infrequent cases, such as cutting food on a bowl with a spoon instead of a knife. Learning embeddings from cooking recipes~\cite{marin2019learning,salvador2017learning} or using language groundings from transcripts~\cite{huang2018finding} could address this issue. On the other hand, we find that explicitly reasoning over objects and human hands provides us a way to extract more information about object properties and the environment. For example, we could observe that people tend to hold the pot to stir, and they usually only handover small objects like a knife instead of a pot. The system is currently designed for cooking tasks, with objects annotated as containers and ingredients. Future work includes generalizing the system to other collaborative tasks as well, such as furniture assembly.
 

More generally, the proposed system generates actions that are executed in an open-loop manner. Future work will consider using the information extracted from online videos to monitor the environment and human state~\cite{chang2019procedure}, optimize sensor placement based on the actions performed in the task~\cite{nikolaidis2009optimal}, adapt robot actions and manage contingencies~\cite{lasota2017survey}.


\noindent\textbf{Implications.} The World Wide Web contains a vast amount of online content that robots can leverage to perform tasks in human-robot and robot-robot teams. We have presented a system that takes as input a previously unseen, cooking video with annotated object labels and outputs a human-interpretable plan. We demonstrate the execution of the plan in a simulation environment with two robotic arms as well as in a real world environment with a human and a robotic arm and show that we can fully reproduce the actions of a simple cooking video. We find that this work brings us closer to the goal of robots executing a variety of manipulation plans by watching cooking videos online. 







\bibliographystyle{IEEEtran}
\footnotesize{
\bibliography{literature}

\begin{thebibliography}{10}
\providecommand{\url}[1]{#1}
\csname url@rmstyle\endcsname
\providecommand{\newblock}{\relax}
\providecommand{\bibinfo}[2]{#2}
\providecommand\BIBentrySTDinterwordspacing{\spaceskip=0pt\relax}
\providecommand\BIBentryALTinterwordstretchfactor{4}
\providecommand\BIBentryALTinterwordspacing{\spaceskip=\fontdimen2\font plus
\BIBentryALTinterwordstretchfactor\fontdimen3\font minus
  \fontdimen4\font\relax}
\providecommand\BIBforeignlanguage[2]{{%
\expandafter\ifx\csname l@#1\endcsname\relax
\typeout{** WARNING: IEEEtran.bst: No hyphenation pattern has been}%
\typeout{** loaded for the language `#1'. Using the pattern for}%
\typeout{** the default language instead.}%
\else
\language=\csname l@#1\endcsname
\fi
#2}}

\bibitem{hoffman2019evaluating}
G.~Hoffman, ``Evaluating fluency in human--robot collaboration,'' \emph{IEEE
  Transactions on Human-Machine Systems}, vol.~49, no.~3, pp. 209--218, 2019.

\bibitem{yang2014cognitive}
Y.~Yang, A.~Guha, C.~Fermuller, and Y.~Aloimonos, ``A cognitive system for
  understanding human manipulation actions,'' \emph{Advances in Cognitive
  Sysytems}, vol.~3, pp. 67--86, 2014.

\bibitem{yang2015robot}
Y.~Yang, Y.~Li, C.~Ferm{\"u}ller, and Y.~Aloimonos, ``Robot learning
  manipulation action plans by "watching" unconstrained videos from the world
  wide web.'' in \emph{AAAI}, 2015, pp. 3686--3693.

\bibitem{wang2021you}
C.-Y. Wang, I.-H. Yeh, and H.-Y.~M. Liao, ``You only learn one representation:
  Unified network for multiple tasks,'' \emph{arXiv preprint arXiv:2105.04206},
  2021.

\bibitem{finn2017one}
C.~Finn, T.~Yu, T.~Zhang, P.~Abbeel, and S.~Levine, ``One-shot visual imitation
  learning via meta-learning,'' in \emph{Conference on robot learning}.\hskip
  1em plus 0.5em minus 0.4em\relax PMLR, 2017, pp. 357--368.

\bibitem{yu2018one}
T.~Yu, P.~Abbeel, S.~Levine, and C.~Finn, ``One-shot hierarchical imitation
  learning of compound visuomotor tasks,'' \emph{arXiv preprint
  arXiv:1810.11043}, 2018.

\bibitem{chen2021learning}
A.~S. Chen, S.~Nair, and C.~Finn, ``Learning generalizable robotic reward
  functions from" in-the-wild" human videos,'' \emph{arXiv preprint
  arXiv:2103.16817}, 2021.

\bibitem{marin2019learning}
J.~Marin, A.~Biswas, F.~Ofli, N.~Hynes, A.~Salvador, Y.~Aytar, I.~Weber, and
  A.~Torralba, ``Recipe1m+: A dataset for learning cross-modal embeddings for
  cooking recipes and food images,'' \emph{{IEEE} Trans. Pattern Anal. Mach.
  Intell.}, 2019.

\bibitem{41880}
\BIBentryALTinterwordspacing
C.~Chelba, T.~Mikolov, M.~Schuster, Q.~Ge, T.~Brants, P.~Koehn, and
  T.~Robinson, ``One billion word benchmark for measuring progress in
  statistical language modeling,'' Google, Tech. Rep., 2013. [Online].
  Available: \url{http://arxiv.org/abs/1312.3005}
\BIBentrySTDinterwordspacing

\bibitem{paulius2018functional}
D.~Paulius, A.~B. Jelodar, and Y.~Sun, ``{Functional Object-Oriented Network:
  Construction \& Expansion},'' in \emph{ICRA}.\hskip 1em plus 0.5em minus
  0.4em\relax IEEE, 2018.

\bibitem{sener2018unsupervised}
F.~Sener and A.~Yao, ``Unsupervised learning and segmentation of complex
  activities from video,'' in \emph{CVPR}, 2018.

\bibitem{lasotabayesian}
P.~A. Lasota and J.~A. Shah, ``Bayesian estimator for partial trajectory
  alignment,'' 2019.

\bibitem{qi2018generalized}
S.~Qi, B.~Jia, and S.-C. Zhu, ``Generalized earley parser: Bridging symbolic
  grammars and sequence data for future prediction,'' \emph{arXiv preprint
  arXiv:1806.03497}, 2018.

\bibitem{zhou2018towards}
L.~Zhou, C.~Xu, and J.~J. Corso, ``Towards automatic learning of procedures
  from web instructional videos,'' in \emph{Thirty-Second AAAI Conference on
  Artificial Intelligence}, 2018.

\bibitem{wang2020boundary}
Z.~Wang, Z.~Gao, L.~Wang, Z.~Li, and G.~Wu, ``Boundary-aware cascade networks
  for temporal action segmentation,'' in \emph{European Conference on Computer
  Vision}.\hskip 1em plus 0.5em minus 0.4em\relax Springer, 2020, pp. 34--51.

\bibitem{li2021temporal}
Z.~Li, Y.~Abu~Farha, and J.~Gall, ``Temporal action segmentation from timestamp
  supervision,'' in \emph{Proceedings of the IEEE/CVF Conference on Computer
  Vision and Pattern Recognition}, 2021, pp. 8365--8374.

\bibitem{huang2020improving}
Y.~Huang, Y.~Sugano, and Y.~Sato, ``Improving action segmentation via
  graph-based temporal reasoning,'' in \emph{Proceedings of the IEEE/CVF
  conference on computer vision and pattern recognition}, 2020, pp.
  14\,024--14\,034.

\bibitem{wang2022sscap}
Z.~Wang, H.~Chen, X.~Li, C.~Liu, Y.~Xiong, J.~Tighe, and C.~Fowlkes, ``Sscap:
  Self-supervised co-occurrence action parsing for unsupervised temporal action
  segmentation,'' in \emph{Proceedings of the IEEE/CVF Winter Conference on
  Applications of Computer Vision}, 2022, pp. 1819--1828.

\bibitem{li2021action}
J.~Li and S.~Todorovic, ``Action shuffle alternating learning for unsupervised
  action segmentation,'' in \emph{Proceedings of the IEEE/CVF Conference on
  Computer Vision and Pattern Recognition}, 2021, pp. 12\,628--12\,636.

\bibitem{hallac2018greedy}
D.~Hallac, P.~Nystrup, and S.~Boyd, ``Greedy gaussian segmentation of
  multivariate time series,'' \emph{Advances in Data Analysis and
  Classification}, pp. 1--25, 2018.

\bibitem{tang2020asynchronous}
J.~Tang, J.~Xia, X.~Mu, B.~Pang, and C.~Lu, ``Asynchronous interaction
  aggregation for action detection,'' in \emph{European Conference on Computer
  Vision}.\hskip 1em plus 0.5em minus 0.4em\relax Springer, 2020, pp. 71--87.

\bibitem{wu2019long}
C.-Y. Wu, C.~Feichtenhofer, H.~Fan, K.~He, P.~Krahenbuhl, and R.~Girshick,
  ``Long-term feature banks for detailed video understanding,'' in
  \emph{Proceedings of the IEEE/CVF Conference on Computer Vision and Pattern
  Recognition}, 2019, pp. 284--293.

\bibitem{tirupattur2021modeling}
P.~Tirupattur, K.~Duarte, Y.~S. Rawat, and M.~Shah, ``Modeling multi-label
  action dependencies for temporal action localization,'' in \emph{Proceedings
  of the IEEE/CVF Conference on Computer Vision and Pattern Recognition}, 2021,
  pp. 1460--1470.

\bibitem{venugopalan2014translating}
S.~Venugopalan, H.~Xu, J.~Donahue, M.~Rohrbach, R.~Mooney, and K.~Saenko,
  ``Translating videos to natural language using deep recurrent neural
  networks,'' \emph{arXiv preprint arXiv:1412.4729}, 2014.

\bibitem{nguyen2018translating}
A.~Nguyen, D.~Kanoulas, L.~Muratore, D.~G. Caldwell, and N.~G. Tsagarakis,
  ``Translating videos to commands for robotic manipulation with deep recurrent
  neural networks,'' in \emph{ICRA}.\hskip 1em plus 0.5em minus 0.4em\relax
  IEEE, 2018.

\bibitem{DBLP:journals/corr/abs-1810-00146}
\BIBentryALTinterwordspacing
H.~Zhang, E.~Heiden, S.~Nikolaidis, J.~J. Lim, and G.~S. Sukhatme,
  ``Auto-conditioned recurrent mixture density networks for learning
  generalizable robot skills,'' \emph{CoRR}, vol. abs/1810.00146, 2018.
  [Online]. Available: \url{http://arxiv.org/abs/1810.00146}
\BIBentrySTDinterwordspacing

\bibitem{sun2018neural}
S.-H. Sun, H.~Noh, S.~Somasundaram, and J.~Lim, ``Neural program synthesis from
  diverse demonstration videos,'' in \emph{ICML}, 2018, pp. 4797--4806.

\bibitem{shu2017learning}
T.~Shu, X.~Gao, M.~S. Ryoo, and S.-C. Zhu, ``Learning social affordance grammar
  from videos: Transferring human interactions to human-robot interactions,''
  \emph{arXiv preprint arXiv:1703.00503}, 2017.

\bibitem{pastra2012minimalist}
K.~Pastra and Y.~Aloimonos, ``The minimalist grammar of action,''
  \emph{Philosophical Transactions of the Royal Society B: Biological
  Sciences}, vol. 367, no. 1585, pp. 103--117, 2012.

\bibitem{chomsky1993lectures}
N.~Chomsky, \emph{Lectures on government and binding: The Pisa lectures}.

\bibitem{hejia_isrr19}
H.~Zhang, P.-J. Lai, S.~Paul, S.~Kothawade, and S.~Nikolaidis, ``Learning
  collaborative action plans from youtube videos,'' in \emph{ISRR 2019}, Hanoi,
  Vietnam, 2019.

\bibitem{Wang_2021_CVPR}
L.~Wang, Z.~Tong, B.~Ji, and G.~Wu, ``Tdn: Temporal difference networks for
  efficient action recognition,'' in \emph{Proceedings of the IEEE/CVF
  Conference on Computer Vision and Pattern Recognition (CVPR)}, June 2021, pp.
  1895--1904.

\bibitem{Chen_2021_ICCV}
Y.~Chen, Z.~Zhang, C.~Yuan, B.~Li, Y.~Deng, and W.~Hu, ``Channel-wise topology
  refinement graph convolution for skeleton-based action recognition,'' in
  \emph{Proceedings of the IEEE/CVF International Conference on Computer Vision
  (ICCV)}, October 2021, pp. 13\,359--13\,368.

\bibitem{cao2018openpose}
Z.~Cao, G.~Hidalgo, T.~Simon, S.-E. Wei, and Y.~Sheikh, ``Open{P}ose: realtime
  multi-person 2{D} pose estimation using {P}art {A}ffinity {F}ields,'' in
  \emph{arXiv preprint arXiv:1812.08008}, 2018.

\bibitem{grundmann2010efficient}
M.~Grundmann, V.~Kwatra, M.~Han, and I.~Essa, ``Efficient hierarchical
  graph-based video segmentation,'' in \emph{CVPR}.\hskip 1em plus 0.5em minus
  0.4em\relax IEEE, 2010, pp. 2141--2148.

\bibitem{jaccard1912distribution}
P.~Jaccard, ``The distribution of the flora in the alpine zone. 1,'' \emph{New
  phytologist}.

\bibitem{church1989stochastic}
K.~W. Church, ``A stochastic parts program and noun phrase parser for
  unrestricted text,'' in \emph{International Conference on Acoustics, Speech,
  and Signal Processing,}, 1989.

\bibitem{holladayforce}
R.~Holladay, T.~Lozano-P{\'e}rez, and A.~Rodriguez, ``Force-and-motion
  constrained planning for tool use,'' \emph{Massachusetts Institute of
  Technology, 2019.}

\bibitem{Berenson:2011:TSR:2046796.2046797}
D.~Berenson, S.~Srinivasa, and J.~Kuffner, ``Task space regions: A framework
  for pose-constrained manipulation planning,'' \emph{Int. J. Rob. Res.}

\bibitem{lavalle2001rapidly}
S.~M. LaValle and J.~J. Kuffner, ``Rapidly-exploring random trees: Progress and
  prospects,'' \emph{Algorithmic and computational robotics: new directions},
  no.~5, pp. 293--308, 2001.

\bibitem{gu2018ava}
C.~Gu, C.~Sun, D.~A. Ross, C.~Vondrick, C.~Pantofaru, Y.~Li,
  S.~Vijayanarasimhan, G.~Toderici, S.~Ricco, R.~Sukthankar, \emph{et~al.},
  ``Ava: A video dataset of spatio-temporally localized atomic visual
  actions,'' in \emph{Proceedings of the IEEE Conference on Computer Vision and
  Pattern Recognition}, 2018, pp. 6047--6056.

\bibitem{carreira2017quo}
J.~Carreira and A.~Zisserman, ``Quo vadis, action recognition? a new model and
  the kinetics dataset,'' in \emph{proceedings of the IEEE Conference on
  Computer Vision and Pattern Recognition}, 2017, pp. 6299--6308.

\bibitem{yang2015grasp}
Y.~Yang, C.~Fermuller, Y.~Li, and Y.~Aloimonos, ``Grasp type revisited: A
  modern perspective on a classical feature for vision,'' in \emph{CVPR}, 2015.

\bibitem{salvador2017learning}
A.~Salvador, N.~Hynes, Y.~Aytar, J.~Marin, F.~Ofli, I.~Weber, and A.~Torralba,
  ``Learning cross-modal embeddings for cooking recipes and food images,'' in
  \emph{CVPR}, 2017.

\bibitem{huang2018finding}
D.-A. Huang, S.~Buch, L.~Dery, A.~Garg, L.~Fei-Fei, and J.~Carlos~Niebles,
  ``Finding it: Weakly-supervised reference-aware visual grounding in
  instructional videos,'' in \emph{CVPR}, 2018.

\bibitem{chang2019procedure}
C.-Y. Chang, D.-A. Huang, D.~Xu, E.~Adeli, L.~Fei-Fei, and J.~C. Niebles,
  ``Procedure planning in instructional videos,'' \emph{arXiv preprint
  arXiv:1907.01172}, 2019.

\bibitem{nikolaidis2009optimal}
S.~Nikolaidis, R.~Ueda, A.~Hayashi, and T.~Arai, ``Optimal camera placement
  considering mobile robot trajectory,'' in \emph{2008 IEEE International
  Conference on Robotics and Biomimetics}.\hskip 1em plus 0.5em minus
  0.4em\relax IEEE, 2009, pp. 1393--1396.

\bibitem{lasota2017survey}
P.~A. Lasota, T.~Fong, J.~A. Shah, \emph{et~al.}, \emph{A survey of methods for
  safe human-robot interaction}.

\end{thebibliography}
}

\end{document}